\documentclass[runningheads]{llncs}
\usepackage{graphicx}
\usepackage{amsmath,amssymb} 
\usepackage{color}

\usepackage{multirow}
\usepackage{booktabs}
\usepackage[width=122mm,left=12mm,paperwidth=146mm,height=193mm,top=12mm,paperheight=217mm]{geometry}

\begin{document}
\title{Pairwise Body-Part Attention for Recognizing Human-Object Interactions}

\titlerunning{Pairwise Body-Part Attention}
%
\author{Hao-Shu Fang\inst{1}\and
Jinkun Cao\inst{1} \and
Yu-Wing Tai\inst{2} \and
Cewu Lu\inst{1}\inst{*}}
\renewcommand{\thefootnote}{\fnsymbol{footnote}}
\footnotetext[1]{The corresponding author is Cewu Lu, email: lucewu@sjtu.edu.cn, twitter: @Cewu\underline{ }Lu, Cewu Lu is a member of MoE Key Lab of Artificial Intelligence, AI Institute, Shanghai Jiao Tong University, and SJTU-SenseTime AI lab.}
%
\authorrunning{H.S. Fang and J. Cao and Y.W. Tai and C. Lu}
%

\institute{Shanghai Jiao Tong University, China \\
\email{fhaoshu@gmail.com, \{caojinkun,lucewu\}@sjtu.edu.cn} \and
Tencent YouTu Lab, China \\
\email{yuwingtai@tencent.com}}
\maketitle              
\begin{abstract}
In human-object interactions (HOI) recognition, conventional methods consider the human body as a whole and pay a uniform attention to the entire body region. They ignore the fact that normally, human interacts with an object by using some parts of the body. In this paper, we argue that different body parts should be paid with different attention in HOI recognition, and the correlations between different body parts should be further considered. This is because our body parts always work collaboratively. We propose a new pairwise body-part attention model which can learn to focus on crucial parts, and their correlations for HOI recognition. A novel attention based feature selection method and a feature representation scheme that can capture pairwise correlations between body parts are introduced in the model. Our proposed approach achieved $\mathbf{10}\%$ relative improvement (36.1 mAP$\rightarrow$ 39.9 mAP) over the state-of-the-art results in HOI recognition on the HICO dataset. We will make our model and source codes \textbf{publicly available}.
\end{abstract}

\section{Introduction}
Recognizing Human-Object Interactions (HOI) in a still image is an important research problem and has applications in image understanding and robotics \cite{aksoy2011learning,worgotter2013simple,yang2013detection}. From a still image, HOI recognition needs to infer the possible interactions between the detected human and objects. Our goal is to evaluate the probabilities of certain interactions on a predefined HOI list.

\begin{figure}[t]
\begin{center}
   \includegraphics[width=1.0\linewidth]{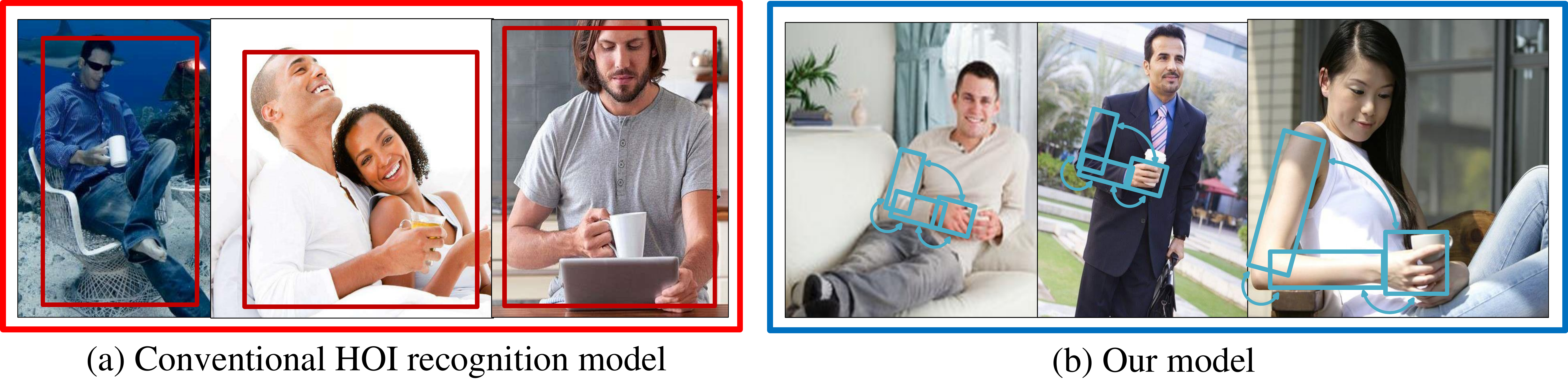}
\end{center}

\caption{Given an image, a person holding a mug in his/her hand, conventional model (a) infers the HOI from the whole body feature. In contrast,  our model  (b) explicitly focuses on discriminative body parts and the correlations between objects and different body parts. In this example, the upper and lower arms which hold a mug form an acute angle across all of the above images.}
\label{fig:intro}

\end{figure}

Conventional methods consider the problem of HOI recognition at holistic body level \cite{thurau2008pose,ikizler2008recognizing,yao2012action} or very coarse part level (e.g., head, torso, and legs) \cite{deepparts} only. However, studies in cognitive science \cite{ro2007attentional,boyer2017attention} have already found that our visual attention is non-uniform, and humans tend to focus on different body parts according to different context. As shown in Figure~\ref{fig:intro}, although the HOI label are the same across all examples, the body gestures are all different except for the arm which holds a mug. This motivates us to introduce a non-uniform attention model which can effectively discover the most informative body parts for HOI recognition.

However, simply building attention on body parts can not capture important HOI semantics, since it ignores the correlations between different body parts. In Figure~\ref{fig:intro}, the upper and lower arms and the hand work collaboratively and form an acute angle due to physical constraints. Such observation motivates us to further focus on the correlations between multiple body parts.  In order to make a practical solution, we consider the joint correlations between each pair of body parts. Such pairwise sets define a new set of correlation feature maps whose features should be extracted simultaneously. Specifically, we introduce pairwise ROI pooling which pools out the joint feature maps of pairwise body parts, and discards the features of other body parts. This representation is robust to irrelevant human gestures and the detected HOI labels have significantly less false positives, since the irrelevant body parts are filtered. With the set of pairwise features, we build an attention model to automatically discover discriminative pairwise correlations of body parts that are meaningful with respect to each HOI label. By minimizing the end-to-end loss, the system is forced to select the most representative pairwise features. In this way, our trained pairwise attention module is able to extract meaningful connections between different body parts.

To the best of our knowledge, our work is the first attempt to apply the attention mechanism to human body part correlations for recognizing human-object interactions.

We evaluate our model on the HICO dataset~\cite{chao2015hico} and the MPII dataset~\cite{andriluka14cvpr}. Our method achieves the state-of-the-art result, and outperforms the previous methods by \textbf{10\%} relatively in mAP on HICO dataset.

\section{Related work}
Our work is related to two active areas in computer vision: human-object interactions and visual attention.
\paragraph{Human-Object Interactions}
Human-object interactions (HOI) recognition is a sub-task of human actions recognition but also a crucial task in understanding the actual human action. It can resolve the ambiguities in action recognition when two persons have almost identical pose and provide a higher level of semantic meaning in the recognition label. Early researches in action recognition consider video inputs. Representative works include~\cite{han2009selection,wang2011action,wang2013action}. In action recognition from still images, previous works attempt to use human pose to recognize human action \cite{wang2006unsupervised,thurau2008pose,ikizler2008recognizing,yang2010recognizing,maji2011action,yao2012action}.

However, considering human pose solely is ambiguous since there is no motion cue in a still image. Human-object interactions are introduced in order to resolve such ambiguities. With additional high level contextual information, it has demonstrated success in improving performance of action recognition~\cite{desai2010discriminative,yao2010modeling,prest2012weakly,hu2013recognising}. Since recognizing the small object is difficult, some works \cite{yao2010grouplet,yao2011combining,sharma2013expanded} attempt to ease the object recognition by recognizing discriminative image patches. Other lines of work include utilizing high level attributes in images~\cite{liu2011recognizing,yao2011human}, exploring the effectiveness of BoF method \cite{delaitre2010recognizing}, incorporating color information \cite{khan2013coloring} and semantic hierarchy \cite{ramanathan2015learning} to assist HOI recognition.

Recently, deep learning based methods~\cite{poseactionrcnn,deepparts,mallya2016learning,gkioxari2017interactnet} give promising results on this task. Specifically, Gkioxari {\it et al.} \cite{deepparts} develop a part based model to make fine-grained action recognition based on the input of both whole-person and part bounding boxes. Mallya and Lazebnik \cite{mallya2016learning} propose a simple network that fuses features from a person bounding box and the whole image to recognize HOIs.

Comparing to the aforementioned methods, especially the deep learning based methods, our method differs mainly in the following aspects. Firstly, our method explicitly considers human body parts and their pairwise correlations, while Gkioxari {\it et al.} \cite{deepparts} only consider parts at a coarse level (i.e., head, torso and legs) and the correlations among them are ignored, and Mallya {\it et al.} \cite{mallya2016learning} only consider bounding boxes of the whole person. Secondly, we propose an attention mechanism to learn to focus on specific parts of body and the spatial configurations, which has not been discussed yet in the previous literatures.

\paragraph{Attention model}
Human perception focuses on parts of the field of view to acquire detailed information and ignore those irrelevant. Such attention mechanism has been studied for a long time in computer vision community. Early works motivated by human perception are saliency detection \cite{itti1998model,hou2007saliency,goferman2012context}. Recently, there have been works that try to incorporate attention mechanism into deep learning framework \cite{mnih2014recurrent,larochelle2010learning,denil2012learning}. Such attempt has been proved to be very effective in many vision tasks including classification \cite{xiao2015application}, detection \cite{ba2014multiple}, image captioning \cite{you2016image,shih2016look,xu2015show} and image-question-answering \cite{yang2016stacked}. Sharma {\it et al.} \cite{sharma2015action} first applied attention model to the area of action recognition by using LSTM~\cite{hochreiter1997long} to focus on important parts of video frames. Several recent works~\cite{liu2017global,song2017end,Girdhar_17b_AttentionalPoolingAction} are partly related to our paper. In~\cite{liu2017global,song2017end}, a LSTM network is used to learn to focus on informative joints of skeleton within each frame to recognize actions in videos. Their method differs from ours that their model learns to focus on discriminative joints of 3D skeleton in an action sequence. In~\cite{Girdhar_17b_AttentionalPoolingAction}, the authors introduce an attention pooling mechanism for action recognition. But their attention is applied to the whole image instead of explicitly focusing on human body parts and the correlations among body parts as we do.

\section{Our Method}
\label{sec:method}
\begin{figure*}[t]
\begin{center}
   \includegraphics[width=0.95\linewidth]{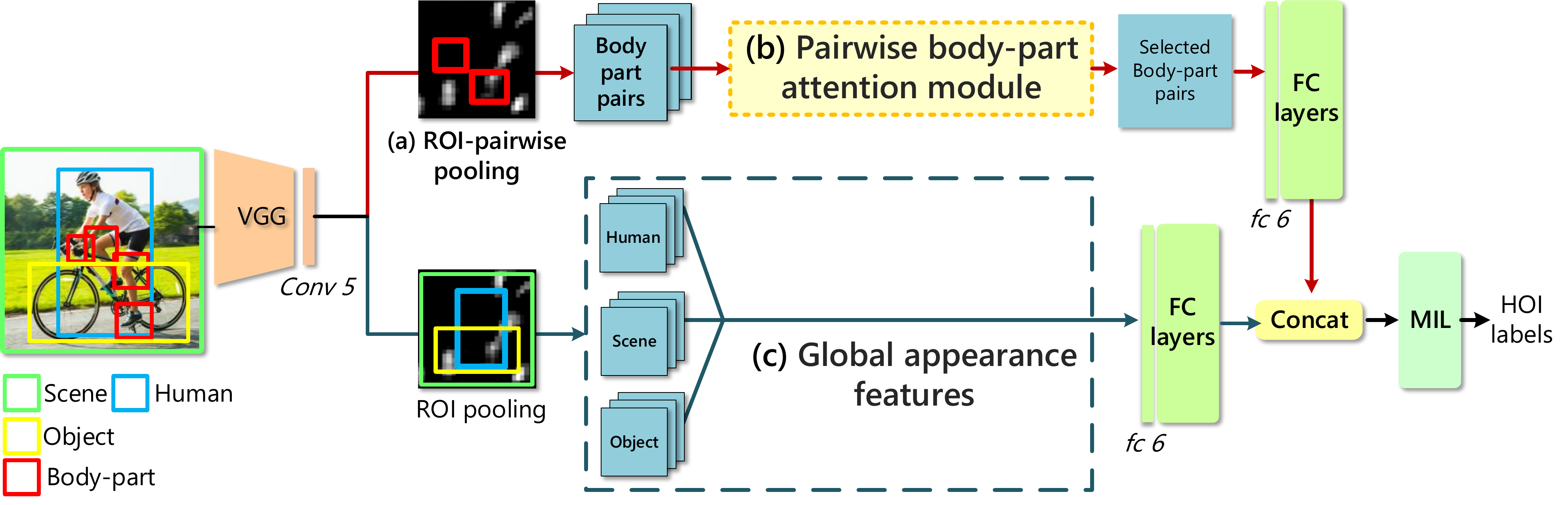}
\end{center}

   \caption{Overview of our framework. The model first extracts visual features of human,object and scene from a set of proposals. We encode the features of different body parts and their pairwise correlations using ROI-pairwise pooling~(a). Then our pairwise body-part attention module (b) will select the feature maps of those discriminative body-part pairs. The global appearance features~(c) from the human, object and scene will also contribute to the final predictions. Following \cite{mallya2016learning}, we adopt MIL to address the problem of multi-person co-occurrence in an image. See text for more details.}
\label{fig:framework}

\end{figure*}

Our approach utilizes both global and local information to infer the HOI labels.

The global contextual information has been well studied by many previous works~\cite{desai2010discriminative,yao2010modeling,prest2012weakly,hu2013recognising}, focusing on utilizing the features of person, object and scene. In section \ref{sec:global}, we review the previous deep learning model~\cite{mallya2016learning} that utilizes features of person and scene. Based on the model from \cite{mallya2016learning}, we further incorporate object features. This forms a powerful base network
which efficiently captures global information. Note that our improved base network has already achieved better performance than the model presented by \cite{mallya2016learning}.

In section \ref{sec:local}, we describe our main algorithm to incorporate pairwise body parts correlations into the deep neural network. Specifically, we propose a simple yet efficient pooling method called ROI-pairwise pooling which encodes both local features of each body part and the pairwise correlations between them. An attention model is developed to focus on discriminative pairwise features. Finally, we present the combination of global features and our local pairwise correlation features in section \ref{sec:com}. Figure~\ref{fig:framework} shows an overview of our network architecture.
\subsection{Global Appearance Features}
\label{sec:global}
\subsubsection{Scene and Human Features}
To utilize the features of the whole person and the scene for HOI recognition, \cite{mallya2016learning} proposed an effective model and we adopt it to build our base network. As shown in Fig.~\ref{fig:framework}, given an input image, we resized and forwarded it through the VGG convolutional layers until the \emph{Conv5} layer. On this shared feature maps, the ROI pooling layer extracts ROI features for each person and the scene given their bounding boxes. For each detected person, the features of him/her are concatenated with the scene features and forwarded through fully connected layers to estimate the scores of each HOI on the predefined list. In the HICO dataset, there can be multiple persons in the same image. Each HOI label is marked as positive as long as the corresponding HOI is observed. To address the issue of multiple persons, the \emph{Multiple Instance Learning}(MIL) framework \cite{maron1998framework} is adopted. The inputs of MIL layer are the predictions for each person in the image, and the output of it is a score array which takes the maximum score of each HOI among all the input predictions. Since MIL is not the major contribution of our work, we refer readers to \cite{mallya2016learning,maron1998framework} for more details of MIL and how it is applied in HOI recognition.

\subsubsection{Incorporating Object Features}
\label{sec:obj}
In order to have a coherent understanding of the HOI in context, we further improve the baseline method by incorporating object features, which is ignored in \cite{mallya2016learning}.

\paragraph{Feature Representation}
Given an object bounding box, a simple solution is to extract the corresponding feature maps and then concatenate them with the existing features of human and scene. However, such method does not have much improvement for the task of HOI recognition. This is because the relative locations between object and human are not encoded. So instead, we set our ROI as a union box of detected human and object. Our experiments (Section \ref{sec:res}) show that such representation is effective.

\paragraph{Handling Multiple Objects}
In HICO dataset, there can be multiple persons and multiple objects in an image. For each person, multiple objects can co-appear around him/her. To solve this problem, we sample multiple union boxes of different objects and the person, and the ROI pooling is applied to each union box respectively. The total number of sampled objects around a person is fixed in our implementation. Implementing details will be explained in Sec. \ref{sec:exp}.

The extracted features of objects are concatenated together with the features of human and scene. This builds a strong base network for capturing well global appearance features.

\subsection{Local Pairwise Body-part Features}
\label{sec:local}
In this subsection, we will describe how to obtain pairwise body-part features using our pairwise body-part attention module.
\subsubsection{ROI-pairwise Pooling}
\begin{figure}[t]
\begin{center}
   \includegraphics[width=0.95\linewidth]{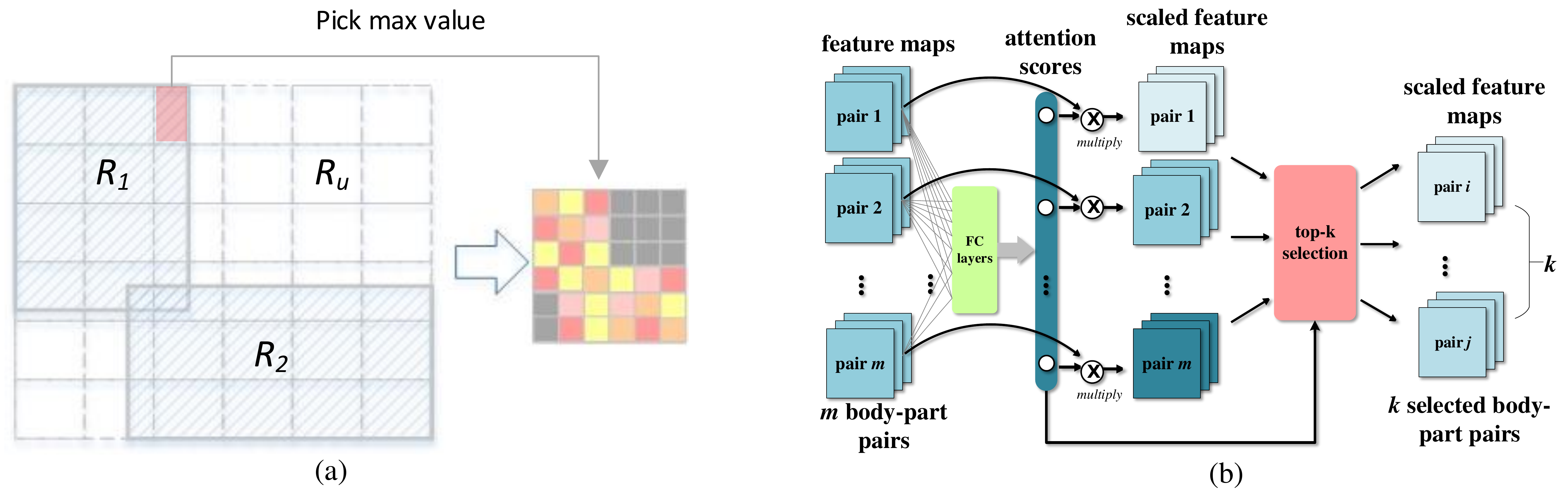}\\
\end{center}

   \caption{ (a) Illustration of the ROI-pairwise pooling layer. The $R_{1}$ and $R_{2}$ each represent a bounding box of different body parts. The ROI-pairwise pooling layer extracts the union area feature of $R_{1}$ and $R_{2}$. The remaining areas are discarded. For each sampled grid location in the ROI-pairwise pooling, the maximum value within the grid area is sampled. (b) Pipeline of the pairwise body-part attention module. From the pairwise body part feature maps pooled by the ROI-pairwise pooling layer, we apply FC layers to estimate the attention score. The attention score is then multiplied with the body part feature maps. Finally, we introduce the feature selection layer which selects the top \emph{k} most important body part pairs and their scaled feature maps are propagated to the next step.}
\label{fig:2}
\end{figure}

Given a pair of body parts, we want to extract their joint feature maps while preserving their relative spatial relationships. Let us denote the ROI pair by $R_{1}(r_{1},c_{1},h_{1},w_{1})$, $R_{2}(r_{2},c_{2},h_{2},w_{2})$, and their union box by $R_{u}(r_{u},c_{u},h_{u},w_{u})$, where $(r,c)$ specifies the top-left corner of the ROI and $(h,w)$ specifies the height and width. An intuitive idea is to set the ROI as the union box of the body-part pair and use ROI pooling layer to extract the features. However, when the two body parts are far from each other, e.g., the wrist and the ankle, their union box would cover a large area of irrelevant body-part. These irrelevant features will confuse the model during training. To avoid it, we assign activation outside (two) body-part boxes as zero to eliminate those irrelevant features. Then, to ensure the uniform size of $R_{u}$ representation, we convert the feature map of union box $R_{u}$ into a fixed size of $H \times W$ feature. It works in a uniformly max-pooling manner:  we first divide the $h_{u} \times w_{u}$ into $H \times W$ grids, then for each grid, the maximum value inside that grid cell is pooled into the corresponding output cell. Figure~\ref{fig:2}(a) illustrates the operation of our ROI-pairwise pooling.

With ROI-pairwise pooling layer, both the joint features of two body parts and their relative location are encoded. Note that the number of body-part pairs are usually big ($C(n,2)$ for $n$  parts) and many pairwise body parts are rarely correlated. We automatically discover those discriminative correlations by proposing an attention module.

\subsubsection{Attention Module}

Figure~\ref{fig:2} (b) illustrates the pipeline of our attention module. Our attention module takes the feature maps of all possible pairwise body-part pairs $P=\{p_{1},p_{2},...,p_{m}\}$ after the ROI-pairwise pooling as input, where $m = C(n,2)$ is the number of body-part pairs. For each pairwise body-part $p_i$, the fully connected layer would regress an attention score $s_i$. The scores $S=\{s_{1},s_{2},...,s_{m}\}$ for $m$ pairwise body-parts indicate the importance of each body-part pair.

\paragraph{Feature Selection} As aforementioned, only some body part pairs are relevant to HOI and irrelevant ones may cause over-fitting of neural network.  Assuming that we need to select features of $k$ body-part pairs, our selection layer will keep the feature maps that belong to the body-part pairs with top-$k$ score and drop the remaining. The selected set can be expressed as:
\begin{equation}\label{eq:select}
    \Phi = \{p_{i} | s_{i} \text{ ranks top } k \text{ in } S\}.
\end{equation}

\paragraph{Attention Allocation}
Different feature maps always have equal value scale, yet they offer different contributions on HOI recognition.  So, we should re-scale the feature maps to reflect their indeed influence. Mathematically, it is modeled as multiplying the corresponding attention score, which can be expressed as:
\begin{equation}\label{eq:score}
    f_{j} = p_{c(j)} \times s_{c(j)},
\end{equation}
where $c(j)$ is the index for the $j^{th}$ element in $\Phi$ and  $f_{j}$ represents the $j^{th}$ re-scaled feature maps.

\paragraph{Discussion} We only allow $k$ pairwise features to represent an interaction. $S$ is forced to assign large value to some pairwise body parts related with input interaction to achieve better accuracy. Therefore, $S$ enables attention mechanism without human supervision. In the experiment section \ref{sec:human_eva}, we verify that the learned attention score is in accord with human perception.

\paragraph{Training}
Since Eqn.~\eqref{eq:select} is not a differentiable function, it has no parameter to be updated and only conveys gradients from the latter layer to the former one during back-propagation. When only the top k pairwise feature maps are selected, the gradients of the feature maps that are selected by the feature selection layer will be copied from latter layer to the former layer. The gradients of the dropped feature maps will be discarded by setting the corresponding values to zero.  Since Eqn.\eqref{eq:score} can be derived easily, the attention scores are updated automatically during back-propagation and our attention module is trained in an end-to-end manner.

Combining the ROI-pairwise pooling layer and the attention module, our pairwise body-part attention module has the following properties:
\begin{itemize}
  \item Both local features of each body part and the higher level spatial relationships between body parts are taken into consideration.
  \item For different HOI, our novel pairwise body-part attention module will automatically discover the discriminative body parts and pairwise relationships.
\end{itemize}

\subsection{Combining Global and Local Features}
\label{sec:com}
After obtaining the selected pairwise body-part features and the global appearance features, we forwarded them through the last FC layers respectively to estimate the final predictions. The prediction is applied for every detected person instances.

\section{Experiment}
\label{sec:exp}
We report our experimental results in this section. We first describe the experimental setting and the details in training our baseline model. Then, we compare our results with those of state-of-the-art methods. Ablation studies are carried to further analyze the effectiveness of each component of our network. Finally, some analyses will be given at the end of this section.
\subsection{Setting}
\subsubsection{Dataset}
We conduct experiments on two frequently used datasets, namely, HICO and MPII dataset. \textbf{HICO dataset}~\cite{chao2015hico} is currently the largest dataset for HOI recognition. It contains 600 HOI labels in total and multiple labels can be simultaneously presented in an image. The ground truth labels are given at image level without any bounding box or location information. Also, multiple persons can appear in the same image, and the activities they perform may or may not be the same. Thus the label can be regarded as an aggregation over all HOI activities in an image. The training set contains 38,116 images and the testing set contains 9,658 images. We randomly sample 10,000 images from the training set as our validation set. \textbf{MPII dataset}~\cite{andriluka14cvpr} contains 15,205 training images and 5708 test images. Unlike HICO dataset, all person instances in an image are assumed to take the same action and each image is classified into only one of 393 action classes. Following~\cite{mallya2016learning}, we sample 6,987 images from the training set as validation set.\\
\subsubsection{HICO}
We use Faster RCNN \cite{ren2015faster} detector to obtain human and object bounding boxes. For each image, 3 human proposals and 4 object proposals will be sampled to fit the GPU memory. If the number of human or objects is less than expected, we pad the remaining area with zero. For the human body parts, we first use pose estimator~\cite{fang2017rmpe} to detect all human keypoints and then define 10 body parts based on keypoints. The selected representative human body parts of our method are shown in Figure~\ref{fig:accuracy-partroi} (a). Each part is defined as a regular bounding box with side length proportional to the size of detected human torso. For body-part pairs, the total number of the pair-wise combination between different body parts is $45(C(10,2))$.

We first try to reproduce Mallya \& Lazebnik~\cite{mallya2016learning}'s result as our baseline. However, with the best of our effort, we can only achieve $35.6$ mAP, while the reported result from Mallya and Lazebnik is 36.1 mAP. We use this model as our baseline model. During training, we follow the same setting as \cite{mallya2016learning}, with an initial learning rate of 1e-5 for 30000 iterations and then 1e-6 for another 30000 iterations. The batch size is set to 10. Similar to the work in~\cite{mallya2016learning,gkioxari2015contextual}, the network is fine-tuned until \emph{conv3} layer. We train our model using Caffe framework~\cite{jia2014caffe} on a single Nvidia 1080 GPU. In the testing period, one forward pass takes 0.15s for an image.

Since the HOI labels in the HICO dataset are highly imbalanced, we adopt a weighted sigmoid cross entropy loss
\begin{equation}\label{eq:loss}
  \text{loss}(I, y) = \sum_{i=1}^{C}  w_p^i \cdot y^i \cdot \log(\hat{y}^i) + w_n^i \cdot (1-y^i) \cdot \log(1-\hat{y}^i),
\nonumber \end{equation}
where $C$ is the number of independent classes, $w_p$ and $w_n$ are weight factors for positive and negative examples, $\hat{y}$ is model's prediction and $y$ is the label for image $I$. Following \cite{mallya2016learning}, we set $w_p=10$ and $w_n=1$.\\
\subsubsection{MPII}
Since all persons in an image are performing the same action, we directly train our model on each person instead of using MIL. The training set of MPII contains manually labeled human keypoints. For testing set, we ran~\cite{fang2017rmpe} to get human keypoints and proposals. The detector~\cite{ren2015faster} is adopted to obtain object bounding boxes in both training and testing sets. Similar to the setting for HICO dataset, we sample a maximum of 4 object proposals per image. During training, we set our initial learning rate as 1e-4, with a decay of 0.1 for every 12000 iterations and stop at 40000 iterations. For MPII dataset, we do not use the weighted loss function for fair comparison with \cite{mallya2016learning}.

\subsection{Results}
\label{sec:res}

\begin{table}
\begin{center}
\begin{tabular}{lcccccr}
\toprule
Method & Full Im. & Bbox/Pose & MIL & Wtd Loss  & mAP \\
\midrule
AlexNet+SVM~\cite{chao2015hico} & \checkmark & & & &19.4 \\
\midrule
R*CNN~\cite{gkioxari2015contextual} & & \checkmark & \checkmark & & 28.5 \\
Mallya \& Lazebnik~\cite{mallya2016learning} & \checkmark & \checkmark & \checkmark & & 33.8 \\
Pose Regu. Attn. Pooling~\cite{Girdhar_17b_AttentionalPoolingAction} & \checkmark & \checkmark & & & 34.6 \\
Ours & \checkmark & \checkmark & \checkmark &  & {\bf 37.5} \\
\midrule
Mallya \& Lazebnik, weighted loss~\cite{mallya2016learning} & \checkmark & \checkmark & \checkmark & \checkmark & 36.1 \\
Ours, weighted loss & \checkmark & \checkmark & \checkmark & \checkmark & {\bf 39.9} \\
\bottomrule
\end{tabular}
\end{center}

  \caption{Comparison with previous results on the HICO test set. The result of R*CNN is directly copied from \cite{mallya2016learning}. }

  \label{tab:HICO}
\end{table}

\begin{table}
\begin{center}
\begin{tabular}{lccccc}
\toprule
Method & Full Img & Bbox & Pose  & Val (mAP) & Test (mAP) \\
\midrule
Dense Trajectory + Pose~\cite{andriluka14cvpr} & \checkmark & & \checkmark & - & 5.5 \\
R*CNN, VGG16~\cite{gkioxari2015contextual} & & \checkmark & & 21.7 & 26.7 \\
Mallya \& Lazebnik, VGG16~\cite{mallya2016learning} & \checkmark & \checkmark & & - & 32.2 \\
Ours, VGG16& \checkmark &\checkmark & \checkmark & {\bf 30.9} & {\bf 36.8} \\
\midrule
Pose Reg. Attn. Pooling, Res101~\cite{Girdhar_17b_AttentionalPoolingAction}& \checkmark & & \checkmark & 30.6 & 36.1 \\
Ours, Res101& \checkmark &\checkmark & \checkmark & {\bf 32.0} & {\bf 37.5} \\
\bottomrule
\end{tabular}
\end{center}

  \caption{Comparison with previous results on the MPII test set. The results on test set are obtained by e-mailing our predictions to the author of~\cite{andriluka14cvpr} }

  \label{tab:MPII}
\end{table}

\begin{figure*}
\begin{center}
   \includegraphics[width=0.9\linewidth]{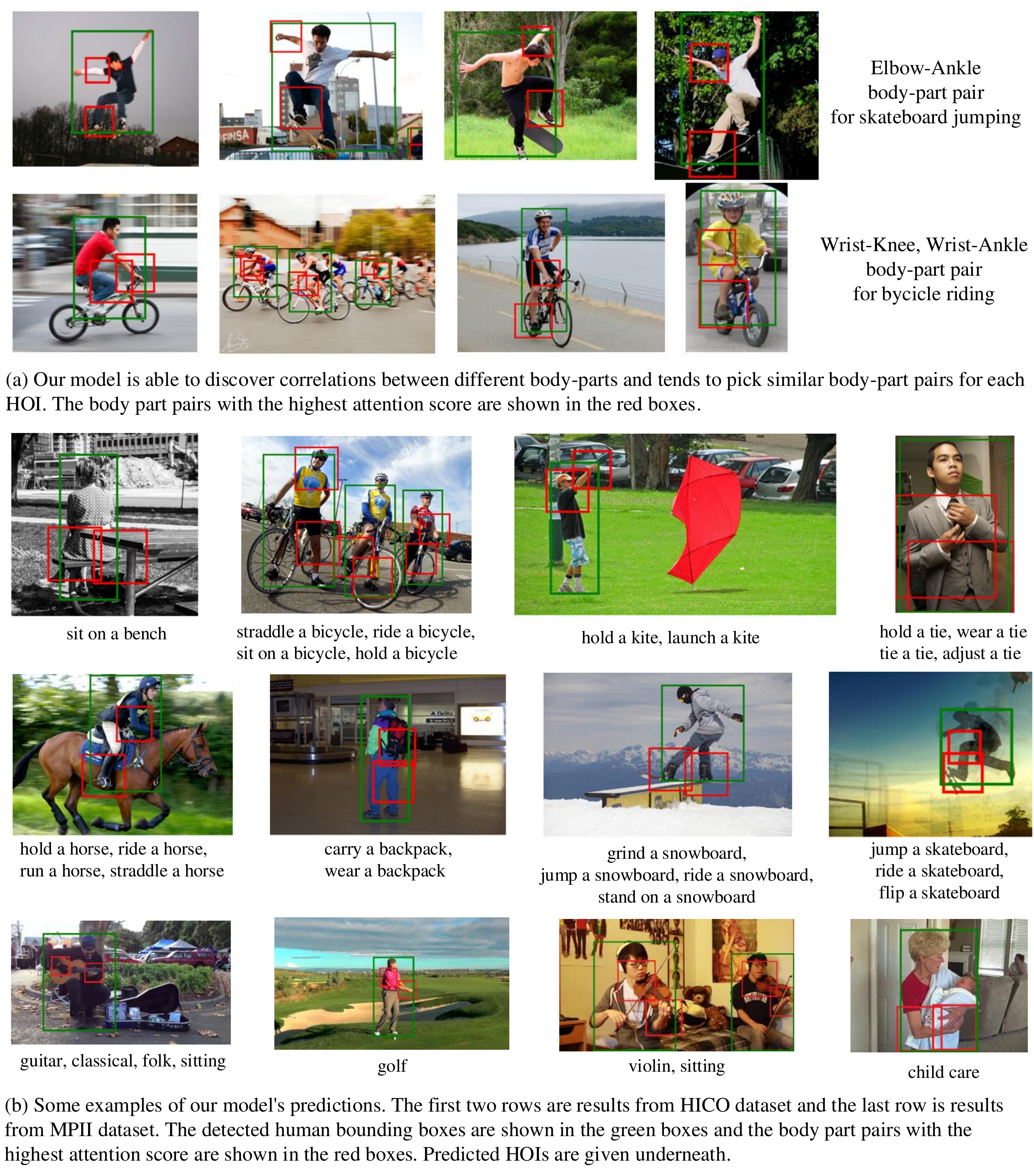}\\
\end{center}

\caption{Results of our model's predictions. }

\label{fig:success}
\end{figure*}
\label{sec:res}
We compare our performance on HICO testing set in Table~\ref{tab:HICO} and on MPII testing set in Table~\ref{tab:MPII}. By selectively focusing on human body parts and their correlations, our VGG16 based model achieves \textbf{37.6} mAP on HICO testing set and \textbf{36.8} mAP on MPII testing set. Using a weighted loss function, we can further achieve \textbf{39.9} mAP on HICO testing set. Since~\cite{Girdhar_17b_AttentionalPoolingAction} use ResNet101~\cite{He2015} as their base model, we also perform an experiment on MPII dataset by replacing our VGG16 base network with the ResNet101 for fair comparison with~\cite{Girdhar_17b_AttentionalPoolingAction}. We can see that our VGG16 based model has already achieved better performance than~\cite{Girdhar_17b_AttentionalPoolingAction} on HICO and MPII dataset, and by using the same base model, we outperform~\cite{Girdhar_17b_AttentionalPoolingAction} by 1.4 mAP on MPII dataset. These results show that the information from body-parts and their correlations is important in recognizing human-object interactions, and it allows us to achieve the state-of-the-art performances on both datasets.

Figure~\ref{fig:success} shows some qualitative results produced by our model. We visualize the body-part pairs with the highest attention score in the red boxes. More results are given in supplementary material.

\subsection{Ablative studies}
To evaluate the effectiveness of each component in our network, we conduct several experiments on HICO dataset and the results are shown in Table~\ref{tab:baselines}.
\begin{table}
  \begin{center}
    \begin{tabular}{ll|c}
      \hline
      & \multicolumn{1}{c|}{\bf Method}  & {\bf mAP }  \\ \hline
      a)& baseline & 35.6 \\ \hline
      \multirow{2}{*}{b)} & union box & {\bf 37.0 } \\
      & tight box & 36.3 \\\hline
      c)& body parts, w/o attention & 38.0 \\\hline
      \multirow{3}{*}{d)} & body-part pairs, w/o attention& 38.9 \\
      & body-part pairs, with attention& {\bf 39.9 }\\
      & body parts \& pairs, with attention& 39.1 \\\hline
    \end{tabular}
  \end{center}

  \caption{Performance of the baseline networks on the HICO test set. ``union box'' refers to the features of an object which are extracted from the area of union box of human and object. ``tight box'' refers to the features of an object which are extracted from the exact area of the object tight box. ``w/o attention'' refers to the method without attention mechanism. }

  \label{tab:baselines}
\end{table}
\subsubsection{Incorporating Object Information}
 As shown in Table~\ref{tab:baselines}(b), our improved baseline model with object features can achieve higher mAP than the baseline method without using object features. It shows that object information is important for HOI recognition. From the table, we can see that using the features from the union box instead of the tight box can achieve higher mAP.  Note that our improved baseline model has already achieved the state-of-the-art results with $0.9$ mAP higher than the results reported by \cite{mallya2016learning}.

\subsubsection{Improvements from body parts information}
We evaluate the performance improvement with additional body-parts information. The feature maps of 10 body parts are directly concatenated with the global appearance features, without taking the advantages of attention mechanism or body-part correlations. As can be seen in Table~\ref{tab:baselines}(c), we further gain an improvement of $1.0$ mAP.

\subsubsection{Pairwise Body-part Attention}
To evaluate the effectiveness of each component of our pairwise body-part attention model, a series of experiments have been carried out and results are reported in Table~\ref{tab:baselines}(d).

Firstly, we consider the correlations of different body parts. The feature maps of the 45 body-part pairs are concatenated with the global appearance features to estimate HOI labels. With body-part pairwise information considered, our model can achieve \textbf{38.9} mAP. It demonstrates that exploiting spatial relationships between body parts benefits the task of HOI recognition.

Then, we add our attention module upon this network. For our feature selection layer, we set $\emph{k}$ as 20. The influence of the value of $k$ will be discussed in the analysis in~\ref{sec:analysis}. With  our pairwise body-part attention model, the performance of our model further yields \textbf{39.9} mAP even though the fully connected layers receive less information from fewer parts.

We also conduct an experiment by simultaneously learning to focus on discriminative body parts and body-part pairs. The candidates for our attention model are the feature maps of 10 body parts and 45 body-part pairs. However, the final result drops slightly to $39.1$ mAP. One possible reason is that our ROI-pairwise pooling has already encoded local features of each single body part. The extra information of body parts may have distracted our attention network.

\subsection{Analysis}
\begin{figure}[t]
\begin{center}
   \includegraphics[width=0.9\linewidth]{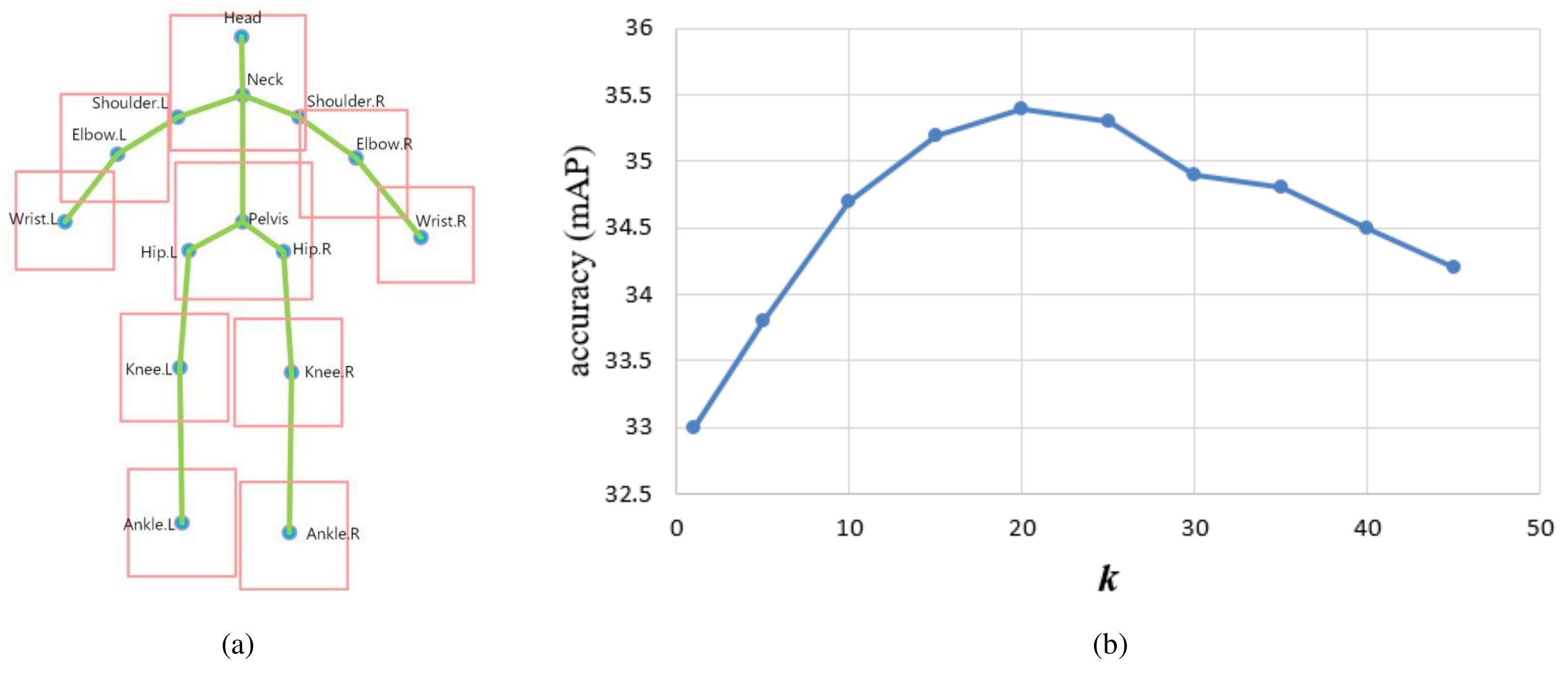}
\end{center}

\caption{(a) Our defined human body-parts. Each bounding box denotes a defined body part. (b) The relationship between recognition accuracy and the number of selected pairwise body part feature maps in the feature selection layer.}
\label{fig:accuracy-partroi}
\end{figure}
\begin{figure*}[t]
\begin{center}
   \includegraphics[width=0.95\linewidth]{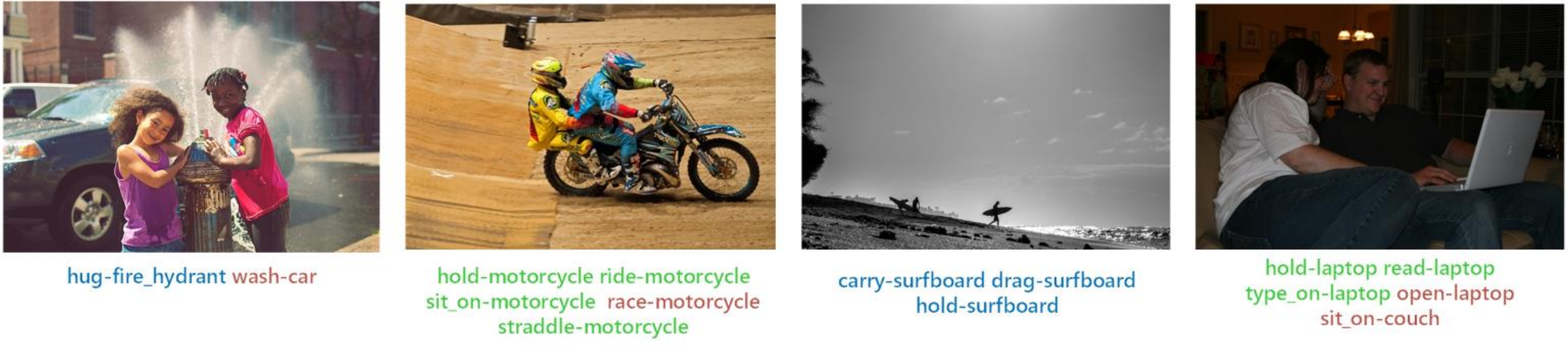}
\end{center}
   \caption{Some failure examples. The false positives of our predictions are labeled in red, and the false negatives of our predictions are labeled in blue. Our model can sometimes get confused with ambiguous background features. Also, when the human or object detectors fail, our approach would fail.}
\label{fig:fail}
\end{figure*}
\label{sec:analysis}
\subsubsection{Parameter for Feature Selection Layer}

In our feature selection layer, we need to decide $k$, the number of body part pairs that we propagate to the next step. We perform an experiment to evaluate the effect of $k$.  We train our pairwise body part attention model on HICO training set with different value of $k$. The performances on validation set are reported in Figure \ref{fig:accuracy-partroi} (b). When $k$ increases, the performance of our model increases until $k=20$. After that, the performance of our model starts to drop. When $k$ equals to 45, it is equivalent to not using the feature selection layer. The performance in this case is 1.2 mAP lower than the highest accuracy. This indicates that rejecting irrelevant body-part pairs is important.

{\small
\begin{table}[t]
  \begin{center}
    \begin{tabular}{l|ccc}
    \hline
     HOI&\multicolumn{3}{c}{Selected correlations}\\
      \hline
      chase-bird& l.knee-r.wrist & r.elbow-neck& r.ankle-r.elbow\\
      board-car & r.ankle-l.elbow& r.ankle-r.elbow& r.elbow-neck\\
      hug-person & l.elbow-neck& r.elbow-neck& r.wrist-neck\\
      jump-bicycle & l.wrist-pelvis & r.ankle-pelvis & r.elbow-neck\\
      adjust-tie & r.wrist-neck & l.wrist-neck & l.elbow-neck\\
      \hline
    \end{tabular}
  \end{center}

  \caption{Some HOIs and their corresponding most selected body-part pairs chosen by our model. The ``l'' and ``r'' flags denote for left and right.}
  \label{tab:examples}

\end{table}
}

\subsubsection{Evaluation of Attention}

\label{sec:human_eva} To see how close the attention of our model is to human's attention, we list out different HOIs and their corresponding body-part pairs that are selected most frequently by our trained attention module. Some examples are presented in Table~\ref{tab:examples}. The entire list is provided in supplementary material. We invite 30 persons to judge whether the choice of the selected pairs are relevant to the given HOI labels. If half of the persons agree that a selected body part pair is important to decide the HOI labels, we regard the selected body part pair as correct. In our setting, the top-\emph{k} accuracy means the correct body part pair appears in the first \emph{k} predictions of attention module. Our top-1 accuracy achieves 0.28 and top-5 accuracy achieves 0.76. It is interesting to see that the body part pairs selected by our attention module match with our intuition to some extent.

\subsubsection{Improvements by HOI class}
To see which kinds of interactions become less confused due to the incorporation of body part information, we compare the results on 20 randomly picked HOIs in HICO dataset with and without the proposed pairwise body-part attention module. The comparisons are summarized in Table~\ref{tab:details}. When the HOIs require more detailed body part information, such as surfboard holding, apple buying and bird releasing, our model shows a great improvement over the baseline model.
{\small
\begin{table}[t]
\begin{center}
    \begin{tabular}{l|c|c|l|c|c}
      \hline
      HOI & \cite{mallya2016learning} & Ours & HOI & \cite{mallya2016learning} & Ours  \\ \hline\hline
      cat scratching & 47.7 & \underline{\textbf{50.9}} & train boarding & 37.1 & \underline{\textbf{48.2}} \\
      umbrella carrying & 83.7 & \underline{\textbf{86.9}} & apple buying & 19.3 & \underline{\textbf{59.0}} \\
      keyboard typing on& \underline{\textbf{71.6}} & 68.3 & cake lighting & 16.3 & \underline{\textbf{24.1}} \\
      boat inspecting & 21.1 & \underline{\textbf{31.9}} & cup inspecting & 1.0 & \underline{\textbf{1.5}} \\
      oven cleaning & \underline{\textbf{22.1}} & 13.1 & fork licking &4.4 & \underline{\textbf{5.3}} \\
      surfboard holding& 52.9 & \underline{\textbf{63.6}} & bird releasing & 14.5 & \underline{\textbf{51.3}} \\
      dining table eating at & 86.6 & \underline{\textbf{86.9}} & car parking & \underline{\textbf{28.9}} & 26.3 \\
      sandwich no interaction & 74.2 & \underline{\textbf{85.2}} & horse jumping & \underline{\textbf{87.0}} & 86.9  \\
      motorcycle washing & 57.7 & \underline{\textbf{64.8}} & spoon washing & 14.5 & \underline{\textbf{15.3}} \\
      airplane loading &\underline{ \textbf{64.1}} & 60.0 & toilet repairing & 11.4 & \underline{\textbf{22.6}} \\
    \end{tabular}
  \end{center}

  \caption{We randomly pick 20 categories in HICO dataset and compare our results with results from Mallya\&Lazebnik~\cite{mallya2016learning}. The evaluation metric is mAP. The full set of results can be found in the supplementary materials.}

  \label{tab:details}
\end{table}
}

\subsubsection{Failure cases}

We present some failure cases in Figure~\ref{fig:fail}. Our model can sometimes get confused with ambiguous background features. For example, the ``wash-car'' label is detected because of the car and water fountain features in the background regardless of the motion of the two persons. Our current approach does not model the interaction of multiple persons to the same object. When two men sitting on the same motorcycle (e.g., the second image), our approach estimates that they are racing motorcycle since there are two detected instances of ``a man riding on a motorcycle''. False-negative can appear in a scene when persons or objects can be hardly detected (e.g., the third image). Our model can also make mistakes when the scene is very ambiguous (e.g., the forth image).

\section{Conclusions}
In this paper, we have proposed a novel pairwise body part attention model which can assign different attention to different body-part pairs. To achieve our goal, we have introduced the ROI pairwise pooling, and the pairwise body-part attention module which extracts useful body part pairs. The pairwise feature maps selected by our attention module are concatenated with background, human, and object features to make the final HOI prediction. Our experimental results show that our approach is robust, and it significantly improves the recognition accuracy especially for the HOI labels which require detailed body part information. In the future, we shall investigate the possibility of including multi-person interactions into the HOI recognition.

\section{Acknowledgement}
This work is supported in part by the National Key R\&D Program of China£¬ No. 2017YFA0700800, National Natural Science Foundation of China under Grants 61772332 and SenseTime Ltd.

\newpage
\bibliographystyle{splncs04}
\bibliography{egbib}

\begin{thebibliography}{10}
\providecommand{\url}[1]{\texttt{#1}}
\providecommand{\urlprefix}{URL }
\providecommand{\doi}[1]{https://doi.org/#1}

\bibitem{aksoy2011learning}
Aksoy, E.E., Abramov, A., D{\"o}rr, J., Ning, K., Dellen, B.,
  W{\"o}rg{\"o}tter, F.: Learning the semantics of object--action relations by
  observation. The International Journal of Robotics Research  \textbf{30}(10),
   1229--1249 (2011)

\bibitem{andriluka14cvpr}
Andriluka, M., Pishchulin, L., Gehler, P., Schiele, B.: 2d human pose
  estimation: New benchmark and state of the art analysis. In: IEEE Conference
  on Computer Vision and Pattern Recognition (CVPR) (June 2014)

\bibitem{ba2014multiple}
Ba, J., Mnih, V., Kavukcuoglu, K.: Multiple object recognition with visual
  attention. In: arXiv preprint arXiv:1412.7755 (2014)

\bibitem{boyer2017attention}
Boyer, T.W., Maouene, J., Sethuraman, N.: Attention to body-parts varies with
  visual preference and verb--effector associations. Cognitive Processing
  (2017)

\bibitem{chao2015hico}
Chao, Y.W., Wang, Z., He, Y., Wang, J., Deng, J.: Hico: A benchmark for
  recognizing human-object interactions in images. In: ICCV (2015)

\bibitem{delaitre2010recognizing}
Delaitre, V., Laptev, I., Sivic, J.: Recognizing human actions in still images:
  a study of bag-of-features and part-based representations. In: BMVC (2010)

\bibitem{denil2012learning}
Denil, M., Bazzani, L., Larochelle, H., de~Freitas, N.: Learning where to
  attend with deep architectures for image tracking. Neural computation
  \textbf{24}(8),  2151--2184 (2012)

\bibitem{desai2010discriminative}
Desai, C., Ramanan, D., Fowlkes, C.: Discriminative models for static
  human-object interactions. In: CVPR'w (2010)

\bibitem{fang2017rmpe}
Fang, H.S., Xie, S., Tai, Y.W., Lu, C.: {RMPE}: Regional multi-person pose
  estimation. In: ICCV (2017)

\bibitem{Girdhar_17b_AttentionalPoolingAction}
Girdhar, R., Ramanan, D.: Attentional pooling for action recognition. In: NIPS
  (2017)

\bibitem{deepparts}
Gkioxari, G., Girshick, R., Malik, J.: Actions and attributes from wholes and
  parts. In: ICCV (2015)

\bibitem{poseactionrcnn}
Gkioxari, G., Hariharan, B., Girshick, R., Malik, J.: R-cnns for pose
  estimation and action detection. In: arXiv preprint arXiv:1406.5212 (2014)

\bibitem{gkioxari2017interactnet}
Gkioxari, G., Girshick, R., Doll\'{a}r, P., He, K.: Detecting and recognizing
  human-object intaractions. In: arXiv preprint arXiv:1704.07333 (2017)

\bibitem{gkioxari2015contextual}
Gkioxari, G., Girshick, R., Malik, J.: Contextual action recognition with r*
  cnn. In: ICCV (2015)

\bibitem{goferman2012context}
Goferman, S., Zelnik-Manor, L., Tal, A.: Context-aware saliency detection.
  TPAMI  \textbf{34}(10),  1915--1926 (2012)

\bibitem{han2009selection}
Han, D., Bo, L., Sminchisescu, C.: Selection and context for action
  recognition. In: ICCV (2009)

\bibitem{He2015}
He, K., Zhang, X., Ren, S., Sun, J.: Deep residual learning for image
  recognition. arXiv preprint arXiv:1512.03385  (2015)

\bibitem{hochreiter1997long}
Hochreiter, S., Schmidhuber, J.: Long short-term memory. Neural computation
  \textbf{9}(8),  1735--1780 (1997)

\bibitem{hou2007saliency}
Hou, X., Zhang, L.: Saliency detection: A spectral residual approach. In: CVPR
  (2007)

\bibitem{hu2013recognising}
Hu, J.F., Zheng, W.S., Lai, J., Gong, S., Xiang, T.: Recognising human-object
  interaction via exemplar based modelling. In: ICCV (2013)

\bibitem{ikizler2008recognizing}
Ikizler, N., Cinbis, R.G., Pehlivan, S., Duygulu, P.: Recognizing actions from
  still images. In: ICPR (2008)

\bibitem{itti1998model}
Itti, L., Koch, C., Niebur, E.: A model of saliency-based visual attention for
  rapid scene analysis. TPAMI  \textbf{20}(11),  1254--1259 (1998)

\bibitem{jia2014caffe}
Jia, Y., Shelhamer, E., Donahue, J., Karayev, S., Long, J., Girshick, R.,
  Guadarrama, S., Darrell, T.: Caffe: Convolutional architecture for fast
  feature embedding. arXiv preprint arXiv:1408.5093  (2014)

\bibitem{khan2013coloring}
Khan, F.S., Anwer, R.M., van~de Weijer, J., Bagdanov, A.D., Lopez, A.M.,
  Felsberg, M.: Coloring action recognition in still images. IJCV
  \textbf{105}(3),  205--221 (2013)

\bibitem{larochelle2010learning}
Larochelle, H., Hinton, G.E.: Learning to combine foveal glimpses with a
  third-order boltzmann machine. In: NIPS (2010)

\bibitem{liu2011recognizing}
Liu, J., Kuipers, B., Savarese, S.: Recognizing human actions by attributes.
  In: CVPR (2011)

\bibitem{liu2017global}
Liu, J., Wang, G., Hu, P., Duan, L.Y., Kot, A.C.: Global context-aware
  attention lstm networks for 3d action recognition. In: CVPR (2017)

\bibitem{maji2011action}
Maji, S., Bourdev, L., Malik, J.: Action recognition from a distributed
  representation of pose and appearance. In: CVPR (2011)

\bibitem{mallya2016learning}
Mallya, A., Lazebnik, S.: Learning models for actions and person-object
  interactions with transfer to question answering. In: ECCV (2016)

\bibitem{maron1998framework}
Maron, O., Lozano-P{\'e}rez, T.: A framework for multiple-instance learning
  (1998)

\bibitem{mnih2014recurrent}
Mnih, V., Heess, N., Graves, A., et~al.: Recurrent models of visual attention.
  In: NIPS (2014)

\bibitem{prest2012weakly}
Prest, A., Schmid, C., Ferrari, V.: Weakly supervised learning of interactions
  between humans and objects. TPAMI  \textbf{34}(3),  601--614 (2012)

\bibitem{ramanathan2015learning}
Ramanathan, V., Li, C., Deng, J., Han, W., Li, Z., Gu, K., Song, Y., Bengio,
  S., Rosenberg, C., Fei-Fei, L.: Learning semantic relationships for better
  action retrieval in images. In: CVPR (2015)

\bibitem{ren2015faster}
Ren, S., He, K., Girshick, R., Sun, J.: Faster r-cnn: Towards real-time object
  detection with region proposal networks. In: NIPS (2015)

\bibitem{ro2007attentional}
Ro, T., Friggel, A., Lavie, N.: Attentional biases for faces and body parts.
  Visual Cognition  \textbf{15}(3),  322--348 (2007)

\bibitem{sharma2013expanded}
Sharma, G., Jurie, F., Schmid, C.: Expanded parts model for human attribute and
  action recognition in still images. In: CVPR (2013)

\bibitem{sharma2015action}
Sharma, S., Kiros, R., Salakhutdinov, R.: Action recognition using visual
  attention (2015)

\bibitem{shih2016look}
Shih, K.J., Singh, S., Hoiem, D.: Where to look: Focus regions for visual
  question answering. In: CVPR (2016)

\bibitem{song2017end}
Song, S., Lan, C., Xing, J., Zeng, W., Liu, J.: An end-to-end spatio-temporal
  attention model for human action recognition from skeleton data. In: AAAI
  (2017)

\bibitem{thurau2008pose}
Thurau, C., Hlav{\'a}c, V.: Pose primitive based human action recognition in
  videos or still images. In: CVPR (2008)

\bibitem{wang2011action}
Wang, H., Kl{\"a}ser, A., Schmid, C., Liu, C.L.: Action recognition by dense
  trajectories. In: CVPR (2011)

\bibitem{wang2013action}
Wang, H., Schmid, C.: Action recognition with improved trajectories. In: ICCV
  (2013)

\bibitem{wang2006unsupervised}
Wang, Y., Jiang, H., Drew, M.S., Li, Z.N., Mori, G.: Unsupervised discovery of
  action classes. In: CVPR (2006)

\bibitem{worgotter2013simple}
W{\"o}rg{\"o}tter, F., Aksoy, E.E., Kr{\"u}ger, N., Piater, J., Ude, A.,
  Tamosiunaite, M.: A simple ontology of manipulation actions based on
  hand-object relations. IEEE Transactions on Autonomous Mental Development
  \textbf{5}(2),  117--134 (2013)

\bibitem{xiao2015application}
Xiao, T., Xu, Y., Yang, K., Zhang, J., Peng, Y., Zhang, Z.: The application of
  two-level attention models in deep convolutional neural network for
  fine-grained image classification. In: CVPR (2015)

\bibitem{xu2015show}
Xu, K., Ba, J., Kiros, R., Cho, K., Courville, A.C., Salakhutdinov, R., Zemel,
  R.S., Bengio, Y.: Show, attend and tell: Neural image caption generation with
  visual attention. In: ICML. vol.~14 (2015)

\bibitem{yang2010recognizing}
Yang, W., Wang, Y., Mori, G.: Recognizing human actions from still images with
  latent poses. In: CVPR (2010)

\bibitem{yang2013detection}
Yang, Y., Fermuller, C., Aloimonos, Y.: Detection of manipulation action
  consequences (mac). In: CVPR (2013)

\bibitem{yang2016stacked}
Yang, Z., He, X., Gao, J., Deng, L., Smola, A.: Stacked attention networks for
  image question answering. In: CVPR (2016)

\bibitem{yao2010grouplet}
Yao, B., Fei-Fei, L.: Grouplet: A structured image representation for
  recognizing human and object interactions. In: CVPR (2010)

\bibitem{yao2010modeling}
Yao, B., Fei-Fei, L.: Modeling mutual context of object and human pose in
  human-object interaction activities. In: CVPR (2010)

\bibitem{yao2012action}
Yao, B., Fei-Fei, L.: Action recognition with exemplar based 2.5 d graph
  matching. In: ECCV (2012)

\bibitem{yao2011human}
Yao, B., Jiang, X., Khosla, A., Lin, A.L., Guibas, L., Fei-Fei, L.: Human
  action recognition by learning bases of action attributes and parts. In: ICCV
  (2011)

\bibitem{yao2011combining}
Yao, B., Khosla, A., Fei-Fei, L.: Combining randomization and discrimination
  for fine-grained image categorization. In: CVPR (2011)

\bibitem{you2016image}
You, Q., Jin, H., Wang, Z., Fang, C., Luo, J.: Image captioning with semantic
  attention. In: CVPR (2016)

\end{thebibliography}
\end{document}